\documentclass[preprint,12pt]{elsarticle}
\usepackage{hyperref}

\hypersetup{
	colorlinks=true,
	linkcolor=cyan,
	filecolor=blue,      
	urlcolor=red,
	citecolor=green,
}
\usepackage{caption} 	
\usepackage{graphicx}	
\usepackage{subfigure}	
\usepackage{float}		
\usepackage{booktabs}	 
\usepackage{diagbox}	 
\usepackage{multirow}	 
\usepackage{makecell} 	 
\usepackage{longtable}	 

\biboptions{numbers,sort&compress}   

\newcommand{\figref}[1]{Fig.\ref{#1}}
\newcommand{\tabref}[1]{Table~\ref{#1}}

\begin{document}
\begin{frontmatter}	
	
\title{MM-SFENet: Multi-scale Multi-task Localization and Classification of Bladder Cancer in MRI with Spatial Feature Encoder Network}		

\author[1,2,3]{Yu Ren \fnref{fn1}}
\author[2,3]{Guoli Wang \fnref{fn1}}	
\author[2,3]{Pingping Wang}		
\author[2,3]{Kunmeng Liu}
\author[1]{Quanjin Liu\corref{cor2}} 
\ead{liuquanjin@aqnu.edu.cn}

\author[4]{Hongfu Sun}
\author[2,3]{Xiang Li}
\author[2,3]{Benzheng Wei\corref{cor2}} 
\ead{wbz99@sina.com}

\address[1]{College of Electronic Engineering and Intelligent Manufacturing, Anqing Normal University, Anqing 246133, China}								
\address[2]{Center for Medical Artificial Intelligence, Shandong University of Traditional Chinese Medicine, Qingdao 266112, China}
\address[3]{Qingdao Academy of Chinese Medical Sciences, Shandong University of Traditional Chinese Medicine, Qingdao 266112, China}
\address[4]{Urological department, Affiliated Hospital of Shandong University of Traditional Chinese Medicine, Jinan 250011, China}

\cortext[cor2]{Corresponding author}			

\fntext[fn1]{Yu Ren and Guoli Wang contributed equally to this work and should be considered co-first authors.}

\begin{abstract}
	\textbf{Background and Objective}: Bladder cancer is a common malignant urinary carcinoma, with muscle-invasive and non-muscle-invasive as its two major subtypes. This paper aims to achieve automated bladder cancer invasiveness localization and classification based on MRI.\\
	\textbf{Method}: Different from previous efforts that segment bladder wall and tumor, we propose a novel end-to-end multi-scale multi-task spatial feature encoder network (MM-SFENet) for locating and classifying bladder cancer, according to the classification criteria of the spatial relationship between the tumor and bladder wall. First, we built a backbone with residual blocks to distinguish bladder wall and tumor; then, a spatial feature encoder is designed to encode the multi-level features of the backbone to learn the criteria. \\
	\textbf{Results}: We substitute Smooth-L1 Loss with IoU Loss for multi-task learning, to improve the accuracy of the classification task. By testing a total of 1287 MRIs collected from 98 patients at the hospital, the mAP and IoU are used as the evaluation metrics. The experimental result could reach 93.34\% and 83.16\% on test set.\\
	\textbf{Conclusions}: The experimental result demonstrates the effectiveness of the proposed MM-SFENet on the localization and classification of bladder cancer. It may provide an effective supplementary diagnosis method for bladder cancer staging.
	
\end{abstract}
\end{frontmatter}	

\section{Introduction}
Bladder Cancer (BC) is considered a critical malignancy with a high recurrence rate. According to the 2022 Global Cancer Statistics report, BC has become the sixth most common cancer worldwide and caused the ninth highest number of deaths \cite{ref1}. Based on the pathological depth of the tumor invasion, BC is characterized as a heterogeneous disease with two major subtypes: muscle-invasive bladder cancer (MIBC) and non-muscle-invasive bladder cancer (NMIBC) \cite{ref2}. Earlier and accurate classification of BC is critical for diagnosis, treatment, and follow-up of oncologic patients \cite{ref3}. Therefore, to improve the efficacy of BC screening and accurate and reliable diagnostic methods are required. 
\par
The current diagnostic methods of BC are great challenges for clinicians, as the available tools for diagnosis and staging include: a) optical cystoscopy (OCy), an invasive and costly method; b) computed tomography (CT); c) magnetic resonance imaging (MRI). OCy is regarded as the gold standard method for BC diagnosis, but this procedure is painful for patients, and it may fail to visualize certain areas within the bladder. Compared with OCy, imaging techniques have been developed to detect tumors non-invasively. Given its high soft-tissue contrast and non-invasive feature, MRI-based image texture analysis technique has made methods using radiomics that predict tumor stage and grade a potential alternative for BC evaluation \cite{ref4}. The popularity of medical imaging equipment has enriched medical image data \cite{ref5}. After training on massive annotated data, data-driven deep learning models are becoming increasingly powerful methods for solving medical imaging problems, such as image reconstruction \cite{ref6}, lesion detection \cite{ref7}, image segmentation \cite{ref8} and image registration \cite{ref9}. In recent years, several studies have shown that deep learning-based tissue segmentation methods using UNet are comparable to segmentation tasks in MRI examinations by radiologists, especially for Bladder Wall (BW) segmentation \cite{ref10}. UNet adopts convolutional neural network (CNN) to downsample and encode the global information of the image, and a short-cut structure is added to fuse the downsampling and upsampling layers by channel splicing \cite{ref11}. This method can well distinguish the texture of BW and tumor well \cite{ref10,ref12,ref13}. Unexpectedly, when reviewing previous literature on deep learning models for BC, we found that exsiting research focuses solely on segmenting the BW using a deep learning model, which has some limitations. So far, there has been no single end-to-end deep neural network that directly localizes and classifies BC. Besides, compared with recently published powerful models for natural images, MRI has a significant untapped opportunity in deep learning-based localization and classification \cite{ref14}. However, BC localization and classification in MRI still have some problems: a) BW and tumor are difficult to delineate because of the very low contrast between them. b) As a tumor invasiveness classification criterion, the spatial relationship between BW and tumor is difficult to learn by a deep-learning model. c) tumors have varied shapes. 
\par
Accordingly, we proposed a systematical model to solve the BC localization and classification problem, namely, Multi-scale Multi-task Spatial Feature Encoder Network (MM-SFENet). Specifically, the anterior half of the proposed network consists of a backbone with residual connection, and a pyramidal spatial feature encoder (SFE) based on feature pyramid networks (FPN) \cite{ref15} to encode different semantic features. The posterior half of the model generates four multi-scale predicted results from different decoders to enhance the classification capability. In terms of the localization task, we substituted the four-variable-independent-regression Smooth-L1 Loss with IoU Loss to improve the performance of the localization. Consequentially, the paper makes three main contributions. 
\begin{enumerate}[-]
	\item We propose a novel end-to-end detector, namely MM-SFENet, to localize and classify BC in MRI, which is the first work of its kind.
	\item We design an encoder SFE considering multi-scale spatial features and embed it into the detector to learn the BC classification criteria. 
	\item Extensive experiments on the datasets indicate the importance of SFE and IoU Loss. Moreover, we conduct comparisons with the latest detectors, and MM-SFENet outperforms state-of-the-art methods.
\end{enumerate}

\section{Related Works}
\textbf{Bladder Segmentation}. With the continuous development of computer hardware, deep learning-based computer-aided diagnosis (CAD) has emerged as an important and fast-growing field in the application of assisting BC segmentation. Recent years have witnessed the extensive application of deep learning-based models for medical image segmentation and detection. Cha et al. \cite{ref16} made a pilot study on bladder CT image segmentation in 2016. In this study, a lesion likelihood map is first generated, and then minor refinement is performed with level sets to obtain the segmented boundaries of the BC. MRI offers manifold advantages over the CT, including its high soft-tissue contrast, and several UNet-based segmentation methods have also been introduced to extract BW in MRI. In 2018, dolz et al. \cite{ref10} first applied UNet to segment BW and tumor in MRI, and they introduced progressive dilation convolutional layers to expand the receptive fields and decrease the sparsity of the dilated kernel. Subsequently, Liu et al. \cite{ref12} take a step further by embedding a pyramidal atrous convolution block into UNet to enlarge receptive fields while capturing multi-scale contextual information, which contributes to accurate segmentation of the BW. Some studies have also been made in 3D CNN. Hammouda et al. \cite{ref13} introduced a 3D framework in T2W MRI to incorporate contextual information of each voxel and then refined the network by a conditional random field. It implies that, with its feature representation ability and high soft-tissue contrast in MRI, CNN can distinguish BC tumor and wall tissue according to their texture. Despite the success of deep learning in BC segmentation, as far as we know, there are few studies regarding BC localization and classification in MRI. Thus this paper conducts a development study on it. 
\par
\textbf{Deep Object Detectors}. The classification of bladder invasive carcinoma subtypes consists of a two-step process: the tumor was first recognized and localized by a bounding box(b-box), then it was further classified by the positional relationships between BW and tumor. 
\par
Contemporary state-of-the-art object detection methods almost follow two major paradigms: two-stage detectors and one-stage detectors. As a standard paradigm of the two-stage detection methods, R-CNN \cite{ref17,ref18,ref19,ref20,ref21,ref22,ref23} architecture combines a proposal detector and a region-wise classifier. In the first stage, Region Proposal Network (RPN) generates a coarse set of region proposals, and in the second stage, region classifiers provide confidence scores and b-box of the proposed region. Selective Search \cite{ref18} was first introduced into region proposals, and R-CNN \cite{ref19} was successfully used in deep learning-based object detection in 2014. To avoid region proposals redundancy, Fast R-CNN \cite{ref20} shares the computations across the proposed regions by RoI pooling, while SPP-net \cite{ref21} extracts regional features by a Spatial Pyramid Pooling (SPP) layer. Despite the success in Fast R-CNN, the region proposals are still generated by traditional methods. Selective Search merges super-pixels based on low-level visual cues rather than a data-driven manner. Region Proposal Network (RPN) is proposed to resolve this issue in a supervised learning way based on CNN. After RPN is embedded into Fast R-CNN, the Faster R-CNN \cite{ref22} is generated. It can replace Selective Search algorithm to form region proposals and reach a trade-off between accuracy and computations. Since the Faster R-CNN is generated, the architecture of two-stage detection methods has been formed and upgraded by its descendants Cascade R-CNN \cite{ref23} and Sparse R-CNN \cite{ref24}, according to multi-stage refinement and relationship between targets, respectively. To better detect objects at multi-scale, an embedded feature pyramid is built to perform feature fusion in FPN that has now become a basic module of the latest detectors due to its excellent performance.
\par
Compared with two-stage detectors, one-stage detectors \cite{ref24,ref25,ref26,ref27,ref28,ref29} are much more desired for real-time object detection, but they are less accurate. The Over-Feat \cite{ref24} model was first introduced as the one-stage method in 2013. SSD \cite{ref25} combines coarse and fine features and makes predictions based on multiple-scale features. RetinaNet \cite{ref26} uses Focal Loss to avoid the extreme foreground-background class imbalance, and first exceeded the two-stage objectors. Later, object detection is considered as a b-box regression in the YOLO model that exhibits a high detection speed. R. Joseph has subsequently made a series of improvements based on YOLO and has proposed v2 and v3 editions \cite{ref27,ref28}, which further focus on increasing detection precision while keeping a very high speed and simple structure. On the original basis of YOLOv3, the architecture of YOLOv4 is optimized comprehensively with good performance \cite{ref29}. These works make significant progress in different areas. The end-to-end BC localization and classification method can be well implemented by a deep learning-based two-stage detector. 

\section{Methods}
\begin{figure*}[htbp]
\vspace{-0.5cm}
\setlength{\abovecaptionskip}{-0.5cm} \setlength{\belowcaptionskip}{-1pt}
\begin{center}
	\includegraphics[width=1\textwidth]{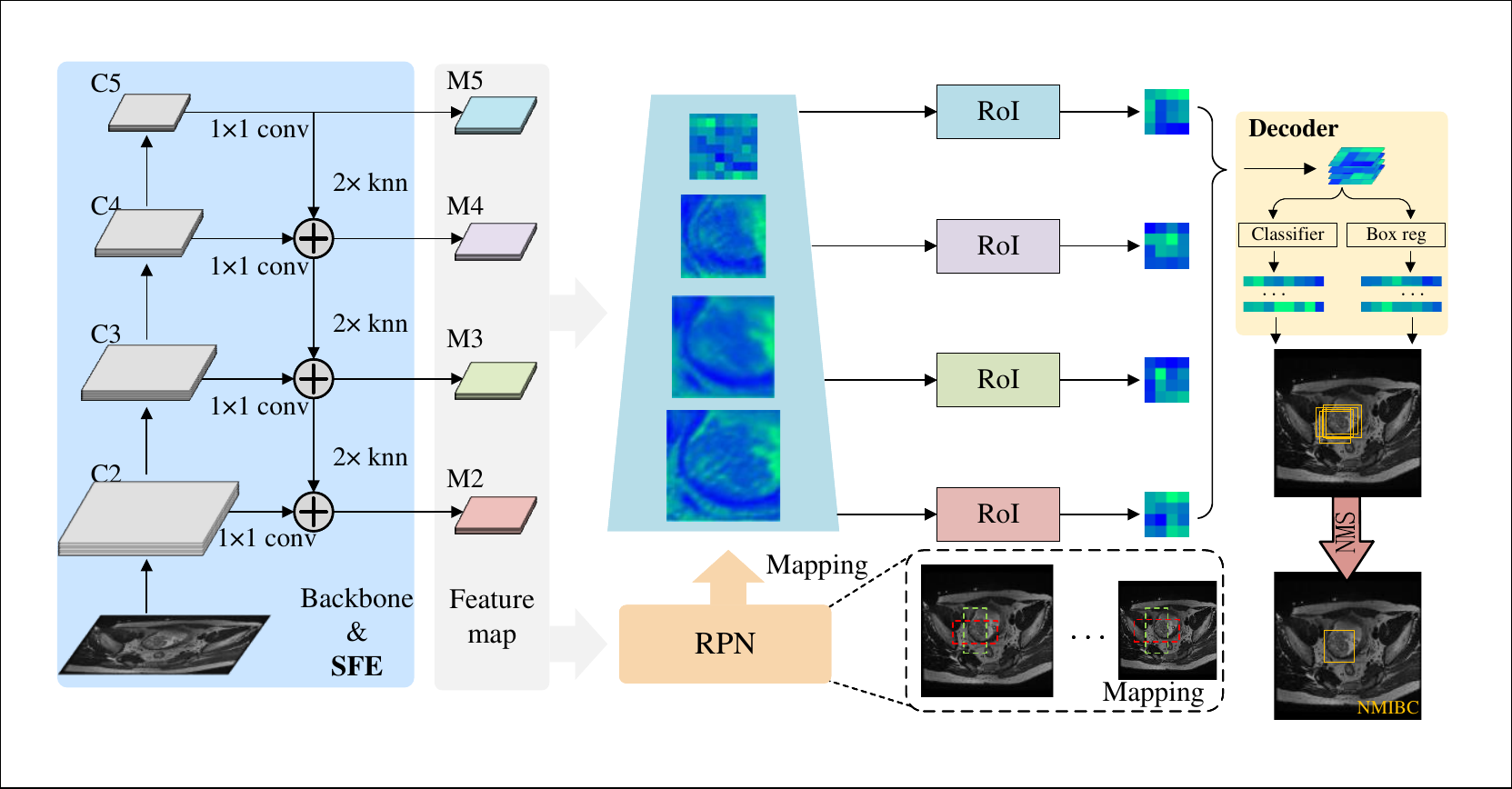}
\end{center}
\caption{Models architecture for detecting and localizing BC. SFE based on FPN make the network can detect at multi-scale and fuse multi-level information at the same time. We assign anchors of a corresponding single scale to multi-level, we define the anchors to have are \{322, 642,1282, 2562\} pixels. Accordingly, RoIs of different scales needed to be assigned to the pyramid levels, and then summarize the outputs to decoder, duplicate predictions are eliminated by NMS operation.}
\label{fig_1}
\end{figure*}

With accurate detection abilities, Faster R-CNN and its descendants have been utilized in medical image analysis with low demand for real-time. Extracting high-level semantic features by constantly downsampling is the first step in R-CNNs, which is achieved by the backbone. Unlike natural images, distinguishing BW from a tumor in a classification task requires a backbone based on a deeper CNN. This information may indicate that how tissue is judged by a CNN with respect to other ones. Standard deep CNNs such as VGG \cite{ref30} have difficulty converging to a minimum and even appear vanishing gradients. The short-cut connection mentioned in the literature as residual connections \cite{ref31}, is used in the backbone to shorten the gradient flow path. Although it has outstanding performance in training deep CNN, backbone built with residual connection presents an important drawback: when the single last feature map is input into RPN to localize suspected lesion areas during inference, the output is only sensitive to a particular size range. To tackle this drawback, one common strategy is adding FPN as a neck after backbone. FPN enables the backbone to not only extracts features in multi-scale but also fuse semantic information from different layers, including the spatial information in lower layers. In addition, the improvement of localization accuracy promotes performance in the classification task. Based on the above discussion, we propose a MM-SFENet, and its detailed architecture is shown in \figref{fig_1}. In this section, the two key technical components of MM-SFENet will be introduced in detail.
\subsection{The Spatial Feature Encoder}
CNN is a structure superimposed by multi-layers of convolutional kernels, and the receptive field size of the network is linearly related to the number of convolutional layers \cite{ref32}. Considering a convolutional kernel in layer l with a size of k and stride s, its receptive field RF can be defined as
\begin{equation}
	R{F_L} = R{F_{L - 1}}{\rm{ + }}\left( {{k_L} - 1} \right) \times \prod\limits_{l = 0}^{L - 1} {{s_l}} 
\end{equation}

From equation (1), it is clear that the receptive field size increase with the number of layers, and each convolutional layer “sees” different information that is an object. 
\begin{figure}[ht]
	\vspace{-0.5cm}
	\setlength{\abovecaptionskip}{-0.7cm} \setlength{\belowcaptionskip}{-0.4cm} 
	\begin{center}
		\includegraphics[width=0.9\textwidth]{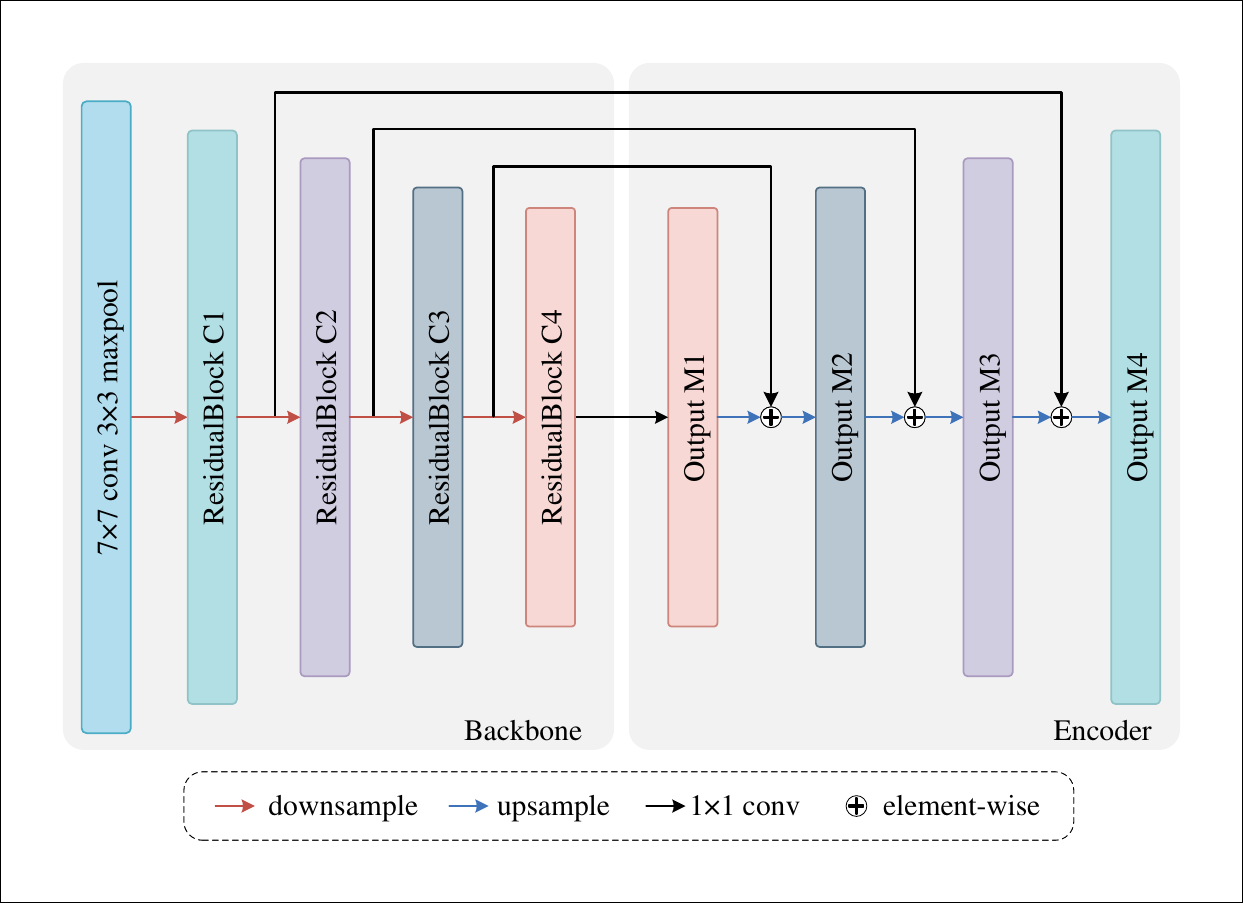}
	\end{center}
	\caption{The architecture of backbone and SFE. SFE taken an arbitrary BC MRI as input, short and long connection inside the network increase the feature representation capability of model.}
	\label{fig_2}
\end{figure}
Therefore, a deep CNN has an inherent multi-scale pyramidal shape in each hierarchy, and higher layers contain more abstract, meaningful and semantically information about the objects, whereas low-level information such as edges, contours, positions and etc. are represented by lower layers. Detectors such as SSD \cite{ref25} and MS-CNN \cite{ref33} detect the objects separately on the feature maps at different backbone hierarchies without mixing high- and low-level information together. This method is called Prediction Pyramid Network (PPN).Despite predicting objects at different multi-scale feature maps, PPN introduces obvious semantic gaps caused by different feature hierarchies, causing it to lack valid semantic information in predicting small objects.Lin et al. \cite{ref27} proposed the FPN to solve the problem. Based on SSD and UNet, FPN constructs a top-down architecture upon the characteristic of CNN by combining multiple shallow and abstract features. We consider the FPN as an encoder that encodes multi-scale features from the backbone and provides multi-level feature representations for the decoder (the detection heads) \cite{ref34}. According to \cite{ref34}, we design a spatial feature encoder SFE to recode the output feature maps and fuse spatial high-resolution features with low-resolution semantic features. The structure diagram is shown in \figref{fig_2}. The construction of SFE involves a backbone, encoder pathway, and lateral connections. 
\par
\begin{figure}[h]
	\vspace{-0.5cm}
	\setlength{\abovecaptionskip}{-0.3cm} \setlength{\belowcaptionskip}{-1pt} 
	\begin{flushleft}
		\includegraphics[width=1\textwidth]{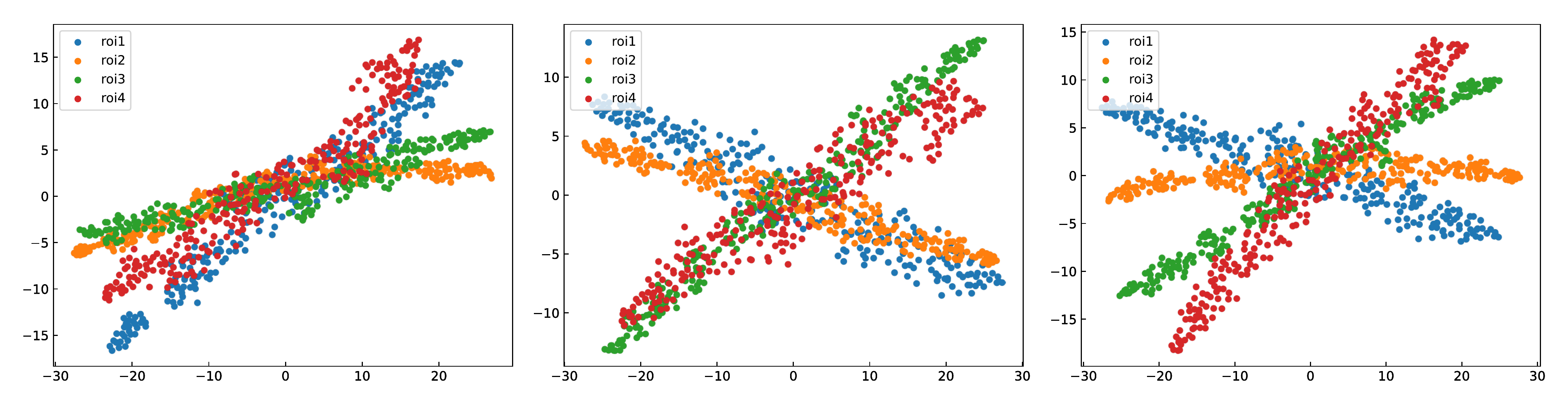}
	\end{flushleft}
	\caption{Depicts the evolution of the feature maps exported by the RoI layers at different training epochs. As the convergence of the network, there are significantly differences between feature maps output by the four RoI layers.}
	\label{fig_3}
\end{figure}

The backbone can be divided into a stem block and four residual blocks \{$C_1$, $C_2$, $C_3$, $C_4$\} under the condition of downsampling magnification \{$2^2$, $2^3$, $2^4$, $2^5$\}. While the encoder outputs the feature maps with a higher spatial resolution using the nearest neighbor upsampling. Then we employ lateral connections to merge feature maps of the same spatial size from the backbone and encoder, which is implemented by an element-wise 1$\times$1 convolutional layer. SFE recodes the feature maps output from different layers in the following way:
\begin{equation}
	{M_1} = L({C_4})
\end{equation}
\begin{equation}
	{M_2} = L({C_3}) \oplus U({M_1})
\end{equation}
\begin{equation}
	{M_3} = L({C_2}) \oplus U({M_2})
\end{equation}
\begin{equation}
	{M_4} = L({C_1}) \oplus U({M_3})
\end{equation}
where $L$ represents lateral connection; $U$ denotes the nearest neighbor upsampling; the final set of feature map is \{$M_1$, $M_2$, $M_3$, $M_4$\} corresponding to \{$C_1$, $C_2$, $C_3$, $C_4$\}. Instead of using the single-level feature, the design of SFE enables us to detect tumors on multi-level features with spatial feature fused, making MM-SFENet learn how to classify the BC invasiveness. \figref{fig_3} is plotted by applying t-SNE \cite{ref35} to the features produced by \{$M_1$, $M_2$, $M_3$, $M_4$\} go through ROI pooling layers.It describes the output feature maps of RoI pooling layer in different training periods. Red, green, yellow, and blue represent the fused feature maps \{$M_4$\}, \{$M_3$, $M_4$\}, \{$M_2$, $M_3$, $M_4$\}, \{$M_1$, $M_2$, $M_3$, $M_4$\}, respectively. It can be seen that there is a difference in the distribution of the multi-level feature maps after being encoded by SFE.
\subsection{IoU Loss}
In the bladder tumor invasiveness classification task, positional relationships between BW and tumor are learned by a classifier to judge the invasiveness, so accurate localization is particularly important. Most detection algorithms use four-variable-independent-regression Smooth-L1 Loss for b-box regression, which is calculated as follows: 
\begin{equation}
{\mathop{\rm Smooth}\nolimits} {\mathop{\rm L}\nolimits} 1Loss(d) = \left\{ {\begin{array}{*{20}{c}}
		{0.5{d^2}}&{\left| d \right| < 1}\\
		{{\rm{  }}\left| d \right| - 0.5}&{{\mathop{\rm otherwise}\nolimits} }
\end{array}} \right.
\end{equation}
Only the Euclidean distance d between points is considered by Smooth L1 Loss, which loses the scale information of the prediction box. It will result in the rapid convergence of the network during the training stages, but the deviation between the prediction box and the ground-truth box will occur during the test stage. IoU Loss \cite{ref36} solves this problem by introducing the intersection and union (IoU) ratio between the prediction box and the ground-truth box. 
\begin{figure}[ht]
	\vspace{-0.5cm}
	\setlength{\abovecaptionskip}{-0.7cm} \setlength{\belowcaptionskip}{-0.4cm} 
	\begin{center}
		\includegraphics[width=0.8\textwidth]{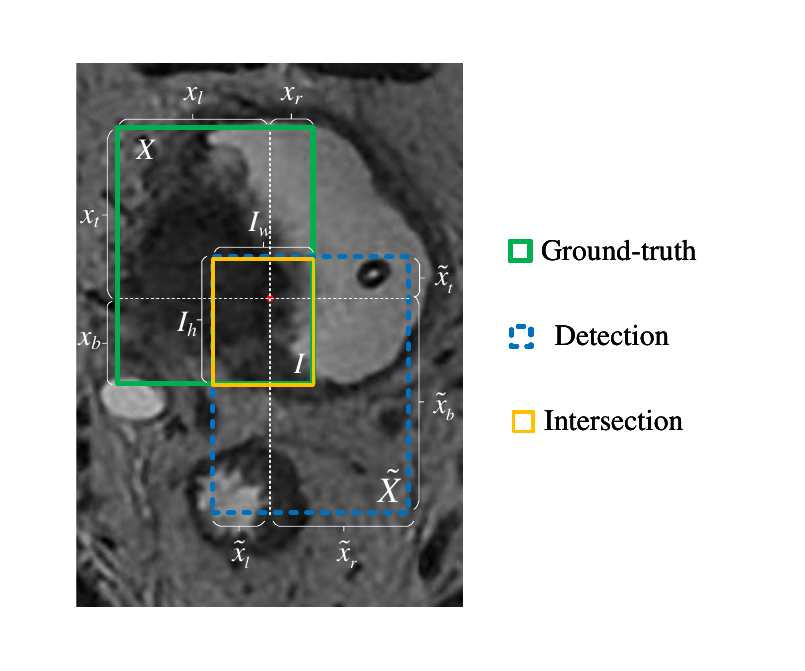}
	\end{center}
	\caption{The architecture of backbone and SFE. SFE taken an arbitrary BC MRI as input, short and long connection inside the network increase the feature representation capability of model.}
	\label{fig_4}
\end{figure}
As shown in \figref{fig_4}, the 4-dimensional vector (${\tilde x_t}$, ${\tilde x_b}$, ${\tilde x_l}$, ${\tilde x_r}$) represents the distance from any pixel in the predication box to the four sides of the predication box, and the ground-truth box is represented as (${x_t}$, ${x_b}$, ${x_l}$, ${x_r}$)
The ground-truth box $X$, the predication box $\tilde X$ and the overlap area $I$ between the ground-truth box and the predication box are calculated as follows: 
\begin{equation}
\left\{ {\begin{array}{*{20}{l}}
		{\tilde X = \left( {{{\tilde x}_t} + {{\tilde x}_b}} \right) * \left( {{{\tilde x}_l} + {{\tilde x}_r}} \right)}\\
		{X = \left( {{x_t} + {x_{\mathop{\rm b}\nolimits} }} \right) * \left( {{x_l} + {x_r}} \right)}\\
		{{I_{\mathop{\rm h}\nolimits} } = \min \left( {{x_t},{{\tilde x}_t}} \right) + \min \left( {{x_{\mathop{\rm b}\nolimits} },{{\tilde x}_b}} \right)}\\
		{{I_{\mathop{\rm w}\nolimits} } = \min \left( {{x_l},{{\tilde x}_l}} \right) + \min \left( {{x_r},{{\tilde x}_{\mathop{\rm r}\nolimits} }} \right)}\\
		{I = {I_w} * {I_h},U = X + \tilde X - I}
\end{array}} \right.
\end{equation}
According to the formula of the IoU ratio, the forward propagation of IoU Loss is calculated as:
\begin{equation}
{L_{loc}} =  - \ln \frac{I}{U}
\end{equation}
About IoU Loss $\tilde x$, the back propagation formula can be obtained by the partial derivative of the following formula: 
\begin{equation}
\frac{{\partial {L_{loc}}}}{{\partial {{\tilde x}_{{\mathop{\rm t}\nolimits} ,b,l,r}}}} = \frac{1}{U}\frac{{\partial \tilde X}}{{\partial {{\tilde x}_{t,b,l,r}}}} - \frac{{U + I}}{{UI}}\frac{{\partial I}}{{\partial {{\tilde x}_{t,b,l,r}}}}
\end{equation}
The loss function is inversely proportional to the intersection area of ${{\partial I} \mathord{\left/{\vphantom {{\partial I} {\partial {{\tilde x}_{t,b,l,r}}}}} \right.\kern-\nulldelimiterspace} {\partial {{\tilde x}_{t,b,l,r}}}}$. The larger the intersection area, the smaller the loss function. The four coordinates of the b-box are regarded as a whole in IoU Loss, which ensures that the predication-box scale is similar to that of the ground-truth box during the b-box regression. From \tabref{tab_1}, it can be seen that IoU Loss acts as a localization loss, enabling better network performance due to the right consistency between IoU Loss and localization evaluation metrics.
\begin{table}[H]
	\centering
	\caption{Comparison between IoU Loss and Smooth L1Loss.}\vspace{-0.3cm} \label{tab_1}
	\begin{tabular}{ccc}
		\toprule	
		Dataset & Loss function & mAP(\%) \\
		\midrule	
		COCO2017 & Smoooth L1 Loss & 36.9 \\
		COCO2017 & IoU Loss & 37.3		  \\
		This paper & Smoooth L1 Loss & 92.92 \\
		This paper & IoU Loss & 93.34		 \\
		\bottomrule	
		
	\end{tabular}
\end{table}
\section{Experiments and Results}
\subsection{Experimental Dataset}
\begin{figure}[htbp]
	\centering
	\vspace{-0.3cm}
	\setlength{\abovecaptionskip}{-0cm} \setlength{\belowcaptionskip}{-1pt} 
	\subfigure[MIBC]{
		\includegraphics[width=0.4\textwidth]{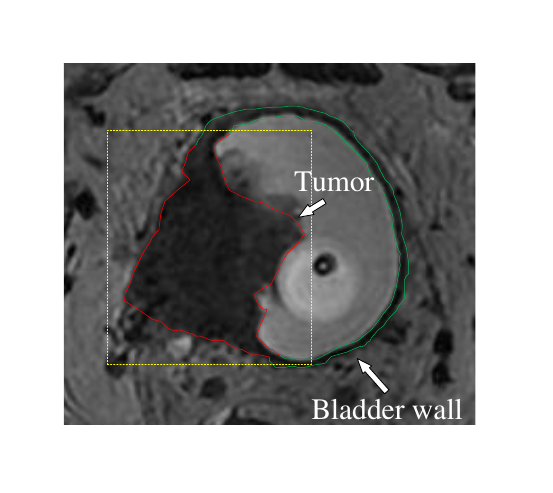} \label{fig_5a}
	}
	\quad
	\subfigure[NMIBC]{
		\includegraphics[width=0.4\textwidth]{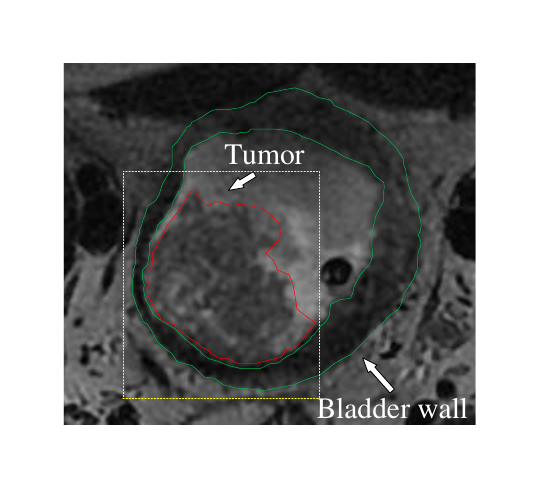} \label{fig_5b} 
	}
	\quad
	\caption{Manual tumor and BW contouring} \label{fig_5}
\end{figure}

MRI images were annotated by professional radiologists from with image-level annotation. BC MRI dataset is constructed by b-box level annotation under the guidance of professional physicians and combined with related materials. The dataset, including data of 98 cases, is divided into MRI images and tumor invasiveness type annotation files. There are 1287 images in the dataset with a uniform resolution of 512$\times$512, of which 130 are used as an independent test set and the rest as a training set. Some images are shown in \figref{fig_5}.
\par
The green lines are BW tissues; the red lines are tumor areas; the yellow b-boxes are labeled boxes. According to whether the tumor invades the BW tissue, \figref{fig_5} can be divided into two types: MIBC tumor of (a) and NMIBC tumor of (b). 
\par
\begin{figure}[htbp]
	\centering
	\vspace{-0.3cm}	
	\setlength{\abovecaptionskip}{-0cm} \setlength{\belowcaptionskip}{-1pt} 
	\subfigure[Distribution of cluster centers]{
		\includegraphics[width=0.45\textwidth]{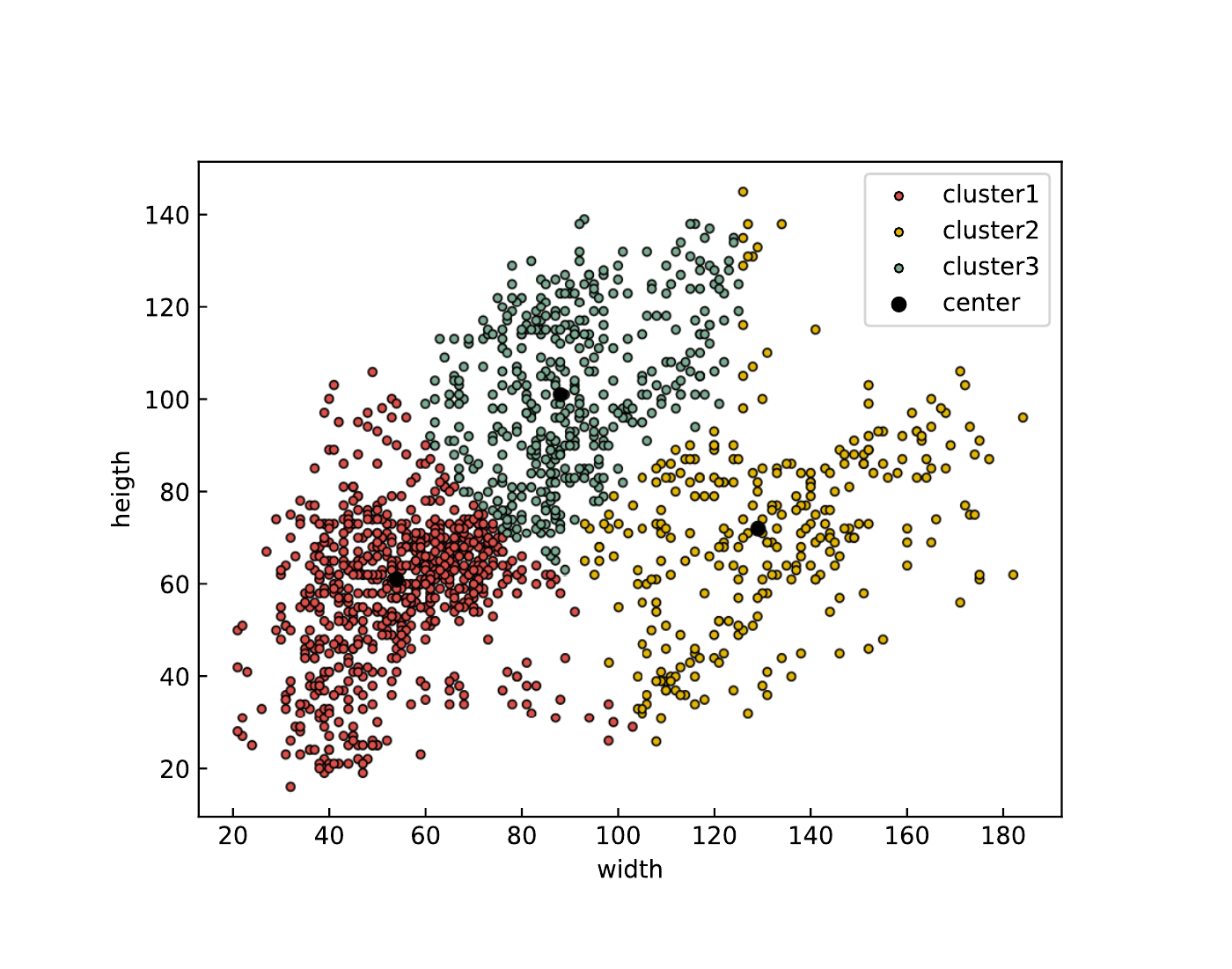} \label{fig_6a}
	}
	\quad
	\subfigure[Histogram of dataset]{
		\includegraphics[width=0.45\textwidth]{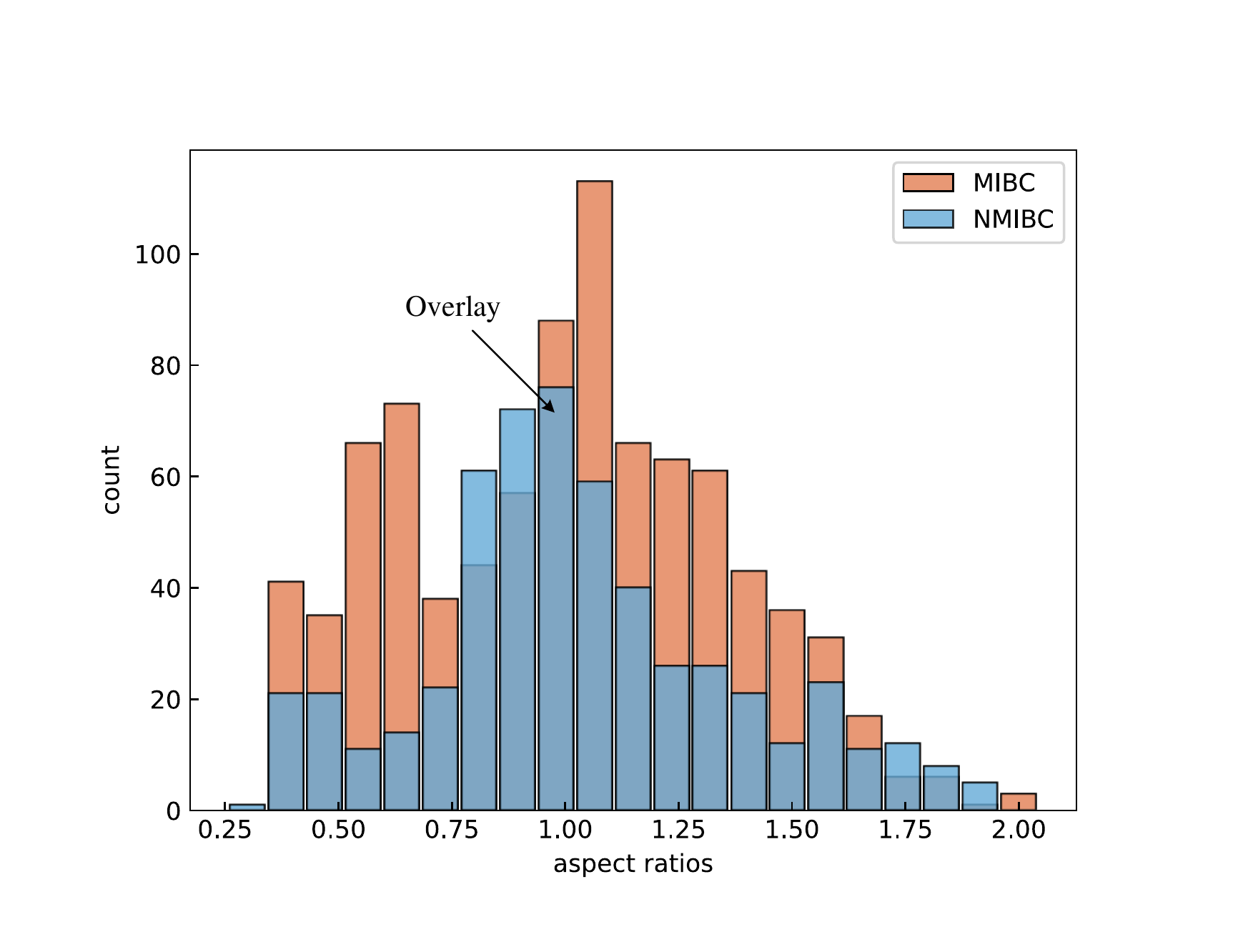} \label{fig_6b} 
	}
	\quad
	\caption{Analysis of dataset} \label{fig_6}
\end{figure}

In anchor-based detectors, matching the anchor box size and shape with the ground-truth box is crucial to the final detection result. The settings of the anchor box in Faster R-CNN are based on the Pascal VOC dataset, which is specific to the natural images but not applicable to the BC MRI dataset. The \textit{k}-means++ is used to cluster the dataset and then regenerate a new setting of anchor box for BC localization. The data clustering results are shown in the \figref{fig_6a}, where the black points represent the cluster centroids, while the other points represent the distribution of the ground-truth box aspects values, and the colors represent the categories to which the ground-truth box belongs. The distribution of the ground-truth box aspect ratio is shown in \figref{fig_6b}.
\par
\figref{fig_6a} shows that the aspect ratio of the ground-truth box is mostly near 0.6, 1.0 and 1.1. Therefore, for the BC MRI dataset, the anchor aspect ratio can be set to [0.6, 1, 1.1], which makes it more consistent with the shape of the true tumor area.
\subsection{Experimental Parameter Settings}
We perform experiments on the deep learning server; the operating system is Ubuntu 18.04, and the CPU is Intel Xeon platinum 8268; The memory size is 32GB; GPU adopts NVIDIA Tesla V100 32GB. All experiments are based on the mmdetection framework. Pytorch version is 1.7.1; mmcv version is 1.3.7, and CUDA version is 11.2. The experimental dataset is divided into two parts on a 9:1: the training set and the test set. During the training process, four processes are started for each graphics card, with 8 images per process. The SGD optimization is used to update the parameters, with the initial learning rate set to 0.02, the weight decay to 0.0001, and the momentum to 0.9. The input image size is uniformly 1000$\times$600, and 24 epochs are trained using a multi-scale training strategy and and adding horizontal and vertical flips as data enhancement, according to \cite{ref37}.
\subsection{Evaluation Metrics and Loss Functions}
AP50 (Average Precision) and mAP (mean Average Precision) are used as the detection evaluation indexes and \textit{IoU} as the localization evaluation index, and AP50 is the \textit{AP} score when IoU=0.5. \textit{AP} is the area under the PR (Precision-Recall) curve . The formula is shown in equation (11). The \textit{mAP }is the mean value of \textit{AP} for each category, and the formula is shown in equation (12). \textit{IoU} is the intersection area of the ground-truth box and the prediction box, and the formula is shown in equation (9),
\begin{equation}
AP = \sum\limits_{k = 1}^N {P(k)\Delta r(k)} 
\end{equation}
where \textit{P}(\textit{k}) is the height of the \textit{k} b-box under the PR curve, and $\Delta r(k)$ is the width of the b-box. The formula for calculating \textit{mAP} is shown as follows,
\begin{equation}
mAP = \frac{{\sum {AP} }}{m}
\end{equation}
where \textit{m} is the total number of categories. The formula for calculating IoU is shown as follows,
\begin{equation}
	IoU\left( {Bo{x_{\mathop{\rm P}\nolimits} },Bo{x_T}} \right) = \frac{{Bo{x_p} \cap Bo{x_T}}}{{Bo{x_{\mathop{\rm p}\nolimits} } \cup Bo{x_T}}}
\end{equation}
where $Box_p$ is the predicted box area, and $Box_T$ is the ground-truth box area
\par
BC detection process can be divided into two parts: localization and classification, based on which we design loss functions for multi-task learning. We use binary cross-entropy for classification and IoU Loss for b-box regression, and the calculation formula is shown as follows (13). \\
\begin{equation}
	\left\{ \begin{array}{l}
		{L_{cls}} = \sum\limits_i { - \log \left( {p_i^*{p_i} + \left( {1 - p_i^*} \right)\left( {1 - {p_i}} \right)} \right)} \\
		{L_{loc}} =  - \ln \left( {IoU} \right)\\
		Loss = \frac{1}{{{N_{pred}}}}{L_{cls}} + \lambda \frac{1}{{{N_{{\mathop{\rm pred}\nolimits} }}}}{L_{loc}}
	\end{array} \right.
\end{equation}
where $p_i^*$ is the probability that the classifier is judged as MIBC, and  $p_i$ is the probability that the classifier is judged as NMIBC. In the total loss function, $N_{pred}$ is the total number of prediction boxes outputted by the final detector, and $\lambda$ is the localization loss function weight. We conduct experiments under the effect of different $\lambda$ on the results, and the experimental results are shown in \tabref{tab_2}. 
\par
\begin{table}[H]
	\centering
	\caption{Detection results with different weights $\lambda$.}\vspace{-0.3cm} \label{tab_2}
	\begin{tabular}{cccc}
		\toprule	
		Dataset & $\lambda=1$ & $\lambda=2$ & $\lambda=5$ \\
		\midrule	
		COCO2017 &36.9 & 37.3 & 35.4 \\
		This paper & \textbf{92.98} & \textbf{93.34} & \textbf{91.56} \\		
		\bottomrule	
	\end{tabular}
\end{table}
As can be seen from \tabref{tab_2}, appropriate $\lambda$ should be targeted for different datasets and tasks. Since the classifier designed in the proposed model needs to classify the tumors and BW based on their position relationship, it is crucial to localize the tumors and BW accurately.
\subsection{Ablation Experiments}
To verify the effectiveness of the improved method proposed in this paper, ablation experiments are conducted on the network model, and the experimental results are shown in \tabref{tab_3}. 
\par
ResNet101 and ResNet50 are used as the backbone in the ablation experiments to test the performance of the proposed model, respectively. As can be seen from \tabref{tab_3}, the SFE increases by 20.53\% and 23.69\% in the networks with ResNet101 and ResNet50 as a backbone, respectively. Using IoU Loss as the loss function, the networks with ResNet101 and ResNet50 as the backbone increase by 2.03\% and 2.89\%, respectively. The experiments prove that the model proposed can still effectively improve the classification and localization accuracy of tumors under different backbones, which also proves the effectiveness of SFE and IoU Loss. 
\begin{table}[H]
	\centering
	\caption{Ablation studies on importance of each components. The results in bold are the best results obtained on the test set.}\vspace{-0.3cm} \label{tab_3}	
	\begin{tabular}{cccccc}
		\toprule	
		Backbone & IoU Loss & SFE & mAP(\%) & AP50(\%) & IoU(\%)  \\
		\midrule	
		ResNet101 &  &  & 71.62 & 71.62 & 63.87 \\
		ResNet101 & $\surd$ &  & 73.65 & 73.63 & 65.98 \\
		ResNet101 &   & $\surd$ & 92.15 & 92.15 & 81.89 \\
		ResNet101 & $\surd$ & $\surd$ & 92.50 & 92.50 & 79.84 \\
		ResNet50 &  &  & 69.23 & 69.20 & 61.89 \\
		ResNet50 & $\surd$ &  & 72.12 & 72.04 & 64.13 \\
		ResNet50 &  & $\surd$ & 92.29 & 92.28 & 81.95 \\
		ResNet50 & $\surd$ & $\surd$ & \textbf{93.34} & \textbf{93.30} & \textbf{83.16} \\
		\bottomrule	
	\end{tabular}
\end{table}
\begin{table}[htbp]
	\caption{Ablation studies on importance of each components. The results in bold are the best results obtained on the test set.}\vspace{-0.3cm} \label{tab_4}
    \scalebox{0.9
    }{
	\centering
	\begin{tabular}{cccccc}
		\toprule	
		Detectors & Backbone & mAP(\%) & AP50(\%) & IoU(\%) & FLOPs(G)  \\
		\midrule	
		Cascade R-CNN & ResNet50 & 92.30 & 92.30 & 82.71 & 155.46 \\
		Sparse R-CNN & ResNet50 & 88.00 & 87.99 & 64.94 & 91.82 \\
		SSD & VGG16 & 92.51 & 92.50 & 81.32 & 207.67 \\
		RetinaNet & Resnet50 & 86.32 & 86.30 & 79.40 & 121.44 \\
		YOLOv4 & CSPDarknet53 & 92.00 & 92.00 & 78.52 & 195.55 \\
		\multirow{5}{*}{MM-SFENet}  &  VGG19 & 90.50 & 90.48  & 78.96  & 261.2 \\
		&  ResNext101 & 89.05 & 89.03  & 77.45 & 266.9 \\
		&  ResNest50 & 80.74  & 80.70  & 75.90  & 173.19 \\
		&  ResNet101 & 92.50  & 92.50  & 79.84  & 266.9 \\
		&  ResNet50 & \textbf{93.34} & \textbf{93.30}  & \textbf{83.16}  & 206.67 \\
		\bottomrule	
	\end{tabular}}
    \end{table}
\subsection{Comparison with the mainstream object detection algorithms}
We propose SFE and substitute the four-variable-independent-regression Smooth-L1 Loss with IoU Loss under Faster R-CNN architecture to localize and classify the invasiveness of BC tumors. The current mainstream detection algorithms are trained on the dataset of this paper, and the obtained results are shown in \tabref{tab_4}, where the highest performance backbone is used for each detector.
\par
Tabel 4 shows that the model proposed showed 93.30\% AP50, 93.34\% mAP, and 83.16\% IoU when using ResNet50 as the backbone. MM-SFENet has better results when compared with the current mainstream detectors, which proves the applicability of the proposed method to the bladder tumor detection problem.
\subsection{Visualization with Smooth Grad-CAM++}
Since the learning mechanism of CNN cannot be fully understood, it is necessary to obtain pathological diagnosis reports based on diagnostic criteria in clinical treatments. Therefore, the class activation heat map is obtained using Smooth Grad-CAM++ \cite{ref38}. The original image size is enlarged by bilinear interpolation and overlaid onto the original image to get the classification visualization image.
\begin{figure}[htbp]
	\centering
	\vspace{-0.3cm}	
	\setlength{\abovecaptionskip}{-0cm} \setlength{\belowcaptionskip}{-1pt} 
	\subfigure[MIBC]{
		\includegraphics[width=0.45\textwidth]{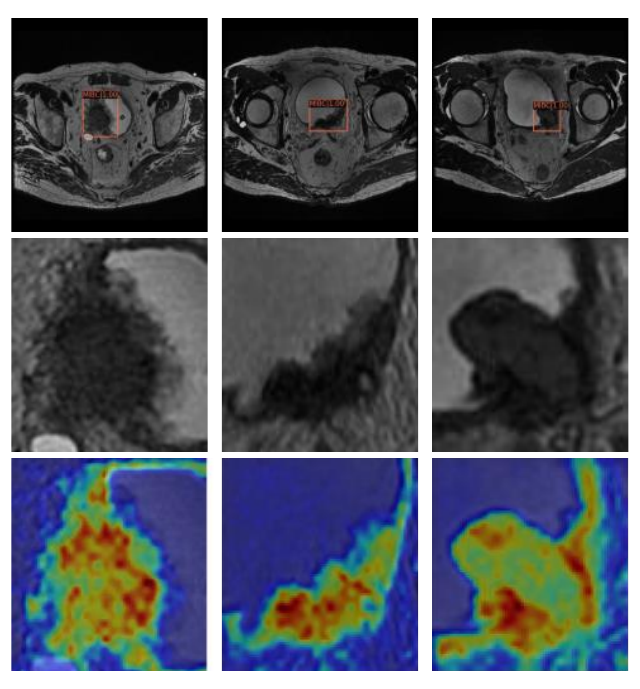} \label{fig_7a}
	}
	\quad
	\subfigure[NMIBC]{
		\includegraphics[width=0.45\textwidth]{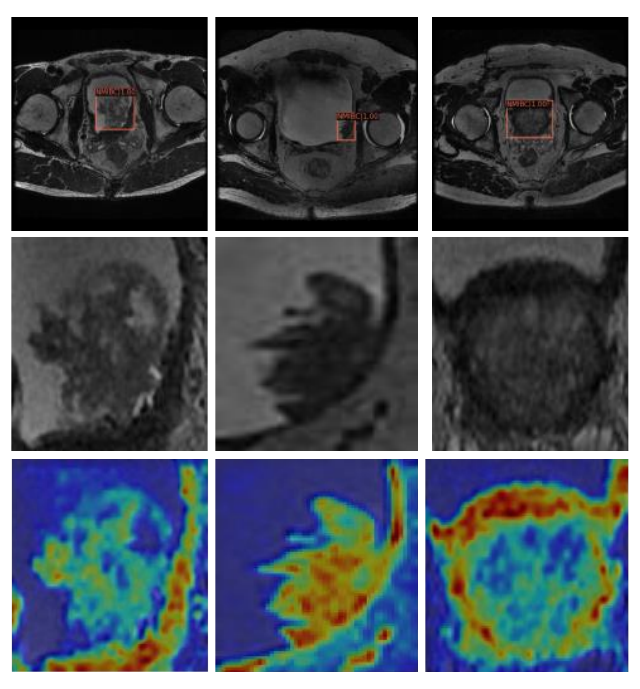} \label{fig_7b} 
	}
	\quad
	\caption{Samples and corresponding heat maps} \label{fig_7}
\end{figure}
The classification heat map can reveal the basis of the model classification. For the determination of NMIBC, the model focuses on the continuity of BW. For the determination of MIBC, the model pays attention to the relationship between the BW and the tumor position, which has significant implications on the interpretability and improvement of the model. 
\section{Discussion}   
All major guidelines agree that Transurethral Resection of Bladder Tumor (TURBT) is the most accurate and reliable technique to acquire a histopathological diagnosis and staging, if performed correctly \cite{ref4}. As a common clinical method for detection and diagnosis of BC, OCy can predict tumor stage with TURBT, but the process is invasiveness and limited field-of-view (FOV) \cite{ref39}. Tumor fragments caused by TURBT often make it difficult for pathologists to judge the invasion depth. With the development of MRI-based texture analysis technology, virtual cystoscopy (VCy) is emerging as a powerful alternative to OCy for the diagnosis of bladder invasive carcinoma subtypes. In the field of BC segmentation, conventional texture analysis methods \cite{ref40,ref41,ref42} clearly perform not well compared with a deep learning-based model, especially for multi-region segmentation. Deep CNN can distinguish the BW and tumor with a more powerful feature extraction capacity, providing the feasibility for the automatic multi-region segmentation in BC MRI images \cite{ref10}. From the analysis of previous literature \cite{ref10,ref12,ref13}, it shows that: a) dilated convolution can expand receptive field while keeping the architecture of models, with more contexture information considered; b) long and short connections are combined in their proposed network, facilitating the gradient flow and alleviating the problem of vanishing gradients; c) capture multi-scale level information at each block.
\par
Despite the impressive performance in BC segmentation, automatic and accurate staging of the BC are not given in previous methods, i.e., MIBC vs NMIBC. Accurate classification of bladder invasive carcinoma subtypes is of vital importance, as each subtype is meant for different treatment modalities \cite{ref2}. In particular, the accurate judgment of NMIBC is the most effective way to accelerate radical treatment and determine the method of preserving intact bladder \cite{ref43}. Therefore, we propose MM-SFENet to fill this gap from an image-level point of view, different from the work by Gabriel \cite{ref44}. Proceeding along with the previous work, backbone built with residual connection and SFE is elaborately designed to approach a multi-scale challenge, as shown in \figref{fig_1}. For the localization task, IoU loss is employed to improve the performance of the classification task by leveraging the correlation between two tasks. From \tabref{tab_3}, we can see the effectiveness of SFE and IoU loss. Benefitting from the excellent tissue contrast of MRI, we can use Grad-CAM to visualize the heat-map of the classification. \figref{fig_7} depicts that which regions will the network pay more attention to, this leads to that BW being used as the classification criteria of MM-SFENet. Furthermore, the proposed method has learned the tumor heterogeneity to make MM-SFENet can distinguish BW and tumor, and then the classification result of BC are given.
\section{Conclusion} 
The proposed MM-SFENet to localize BC and classify their invasiveness in MRI, according to the position relationship between the tumor and the BW. The model uses SFE to fuse spatial high-resolution features with low-resolution semantic features and output four multi-scale feature maps to detect tumors. In the localization task, IoU Loss is used to substituting Smooth L1 Loss to improve the tumor localization accuracy, and the settings of the anchor-box are reconstructed using k-means++. The BC MRI dataset is constructed with instructions from professional physicians. The experiments showed the effectiveness of MM-SFENet in localizing and classifying BC. The automatic BC staging technique should be further studied for the further staging of the BC based on the invasiveness classification and help doctors in formulating treatment plans.
\section{Declaration of Competing Interest}  
The authors declare no conflicts of interest.
\section{Acknowledgements}
This work is partly supported by the National Nature Science Foundation of China (No.61872225), the Natural Science Foundation of Shandong Province (No.ZR2020KF013, No.ZR2020ZD44, No.ZR2019ZD04, No. ZR2020QF043) and Introduction and Cultivation Program for Young Creative Talents in Colleges and Universities of Shandong Province (No.2019-173), the Special fund of Qilu Health and Health Leading Talents Training Project.
\end{document}